\documentclass[letterpaper]{article} 
\usepackage{aaai2026}  
\usepackage{times}  
\usepackage{helvet}  
\usepackage{courier}  
\usepackage[hyphens]{url}  
\usepackage{graphicx} 
\usepackage{natbib}  
\usepackage{caption} 
\usepackage{algorithm}
\usepackage{algorithmic}

\usepackage{amsmath}
\usepackage{amsfonts}
\usepackage{graphicx}
\usepackage{multirow}  
\usepackage{booktabs}  
\usepackage{siunitx} 
\usepackage{microtype} 
\usepackage{booktabs}
\usepackage{subcaption}
\usepackage{newfloat}
\usepackage{listings}
\DeclareCaptionStyle{ruled}{labelfont=normalfont,labelsep=colon,strut=off} 
\floatstyle{ruled}
\newfloat{listing}{tb}{lst}{}
\floatname{listing}{Listing}
\title{Deep Hidden Cognition Facilitates
Reliable Chain-of-Thought Reasoning}
\author{
    Zijun Chen,
    Wenbo Hu\thanks{Corresponding Author.},
    Richang Hong
}
\affiliations{
    Hefei University of Technology\\
    chenzijun@mail.hfut.edu.cn, \{wenbohu,hongrc\}@hfut.edu.cn
}

\usepackage{bibentry}

\begin{document}

\maketitle

\begin{abstract}
Chain of Thought (CoT) reasoning has demonstrated remarkable deep reasoning capabilities in both large language models (LLMs) and multimodal large language models (MLLMs). However, its reliability is often undermined by the accumulation of errors in intermediate steps. This paper proposes a novel approach to calibrating CoT reasoning accuracy by leveraging the model’s internal cognition of truthfulness. Our findings suggest that the model implicitly tracks the evolving veracity of intermediate steps throughout the dynamic, progressive reasoning process. We train a confidence predictor to quantify the model’s internal cognition of truthfulness at each reasoning step, enabling dynamic selection of the most plausible reasoning path through beam search. Experimental results demonstrate that our method significantly outperforms the state-of-the-art baselines (e.g., Self-Consistency, and PRM Guided Search) across the mathematical, symbolic, and commonsense reasoning tasks, exhibiting superior accuracy and reliability in both unimodal and multimodal settings. This study proposes a novel path toward improving the reliability of CoT reasoning, demonstrating strong potential for wide-ranging applications.
\end{abstract}

\begin{links}
    \link{Code}{https://github.com/hfutml/cog-cot}
\end{links}

\section{Introduction}
Chain of Thought (CoT) reasoning has emerged as a powerful paradigm for unlocking deep reasoning capabilities in large language models (LLMs, \cite{wei2022chain,lyu2023faithful}) and multimodal large language models (MLLMs, \cite{zhang2023multimodal,zheng2023ddcot}). CoT not only enhances the model's performance but also ensures that the model has greater adaptability and controllability in complex, dynamic real-world tasks.


\begin{figure}[t]  
    \centering
    \includegraphics[width=1\columnwidth]{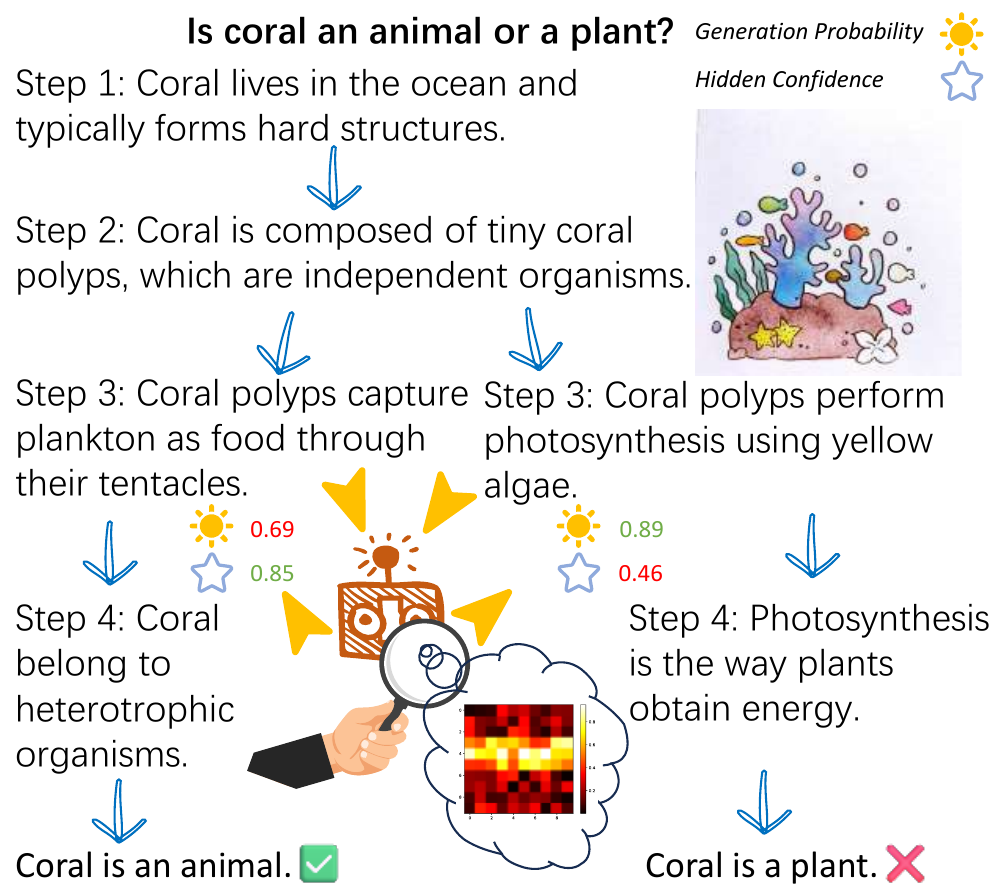}
    \caption{Demonstrate of the dissociation between surface-level generation probabilities and latent cognition in reasoning chains. The internal activations contain hidden representations of true information, and these sensitive activations are primarily concentrated in the intermediate layers.}
    \label{fig:display}
\end{figure}

Yet, the CoT approach is not without its Achilles’ heel. Evidence suggests that CoT does not consistently outperform direct question-answering methods \cite{sprague2024cot}, a limitation rooted in the progressive accumulation of errors. When an intermediate step contains an error, it can lead to the final result being incorrect. A key issue in mitigating this phenomenon is identifying potential errors generated by the model during the generation process and then correcting them. Some researchers have explored this problem: \citet{xie2023self} utilized large models for self-evaluation, taking the probability of the model assigning items to the True category as an indicator of the correctness of intermediate steps. \citet{yin2024reasoning} identified errors by detecting abrupt changes in the negative log probability values of tokens during the model's generation process. Self-consistency \cite{wang2022self} selects candidates with high voting rates or generation probabilities from multiple samplings as decoding results. Although such methods have achieved certain effects, they all rely solely on the surface probabilities of tokens generated by the model. However, the overconfidence of large models has been widely confirmed \cite{kadavath2022language,tian2023just,chen-etal-2025-unveiling}, and their generation probabilities do not match the actual truth, which inevitably raises concerns about the reliability of similar methods.

A profound paradox of human nature lies in the frequent disparity between what is spoken and what is thought. Intriguingly, this duality mirrors latent mechanisms in LLMs: even when generating erroneous outputs, their pretrained latent spaces retain structured factual knowledge, as evidenced by neural activation patterns that encode verifiable information \cite{burns2022discovering,moschella2022relative,rai2024investigation,skean2025layer}.

Building on these studies, we consider whether the internal states of the model during CoT reasoning can generate exploitable authenticity signals. Such signals do not rely on surface-level generation probabilities and can assign appropriate probabilities even to incorrect outputs. These authenticity signals can then guide more reliable CoT decoding. \citet{liu2024enhancing} used the hidden states of the model's final layer to train a confidence predictor, achieving higher calibration than generation probabilities. However, the hidden states of the model's final layer have been condensed into the probability prediction of the next token, and the authenticity signals encoded therein may not be the most dense.
Studies have demonstrated that large models handle information differently across layers: lower layers focus on basic semantics, while higher layers process more abstract knowledge \cite{tenney2019bert,dai2021knowledge,meng2022locating}. Research by \citet{li2023inference} shows that the encoding of such authenticity is dispersed and primarily concentrated in the middle layers. Therefore, extracting these dispersed authenticity signals may be more beneficial for quantifying authenticity probabilities compared to using only the final layer.

In this paper, we propose a novel method: using probing techniques to identify attention heads most sensitive to truthfulness, extracting their activations to construct a multi-layered feature representation, and training a high-precision confidence predictor. This predictor can accurately assess the correctness of each reasoning step. Furthermore, we integrate this predictor into a step-by-step decomposed CoT framework, combining it with beam search to dynamically select high-confidence reasoning steps. This strategy effectively suppresses error accumulation and enhances the overall reliability of CoT.
This work advances the field through three key contributions:
\begin{enumerate}
    \item we identified that a model's internal activations during CoT reasoning encode truthfulness information, enabling us to train a confidence predictor with exceptional calibration performance;
    \item we developed a binary-labeled CoT dataset, which supports training the predictor and fosters research into CoT reliability; this dataset will be made publicly available;
    \item we introduced a novel method that leverages the model's internal cognitive verification to enhance the reliability of CoT generation, with its effectiveness demonstrated across diverse unimodal and multimodal tasks and various models, including the large reasoning model.
\end{enumerate}

\begin{figure*}[t]  
    \centering
    \includegraphics[width=2.1\columnwidth]{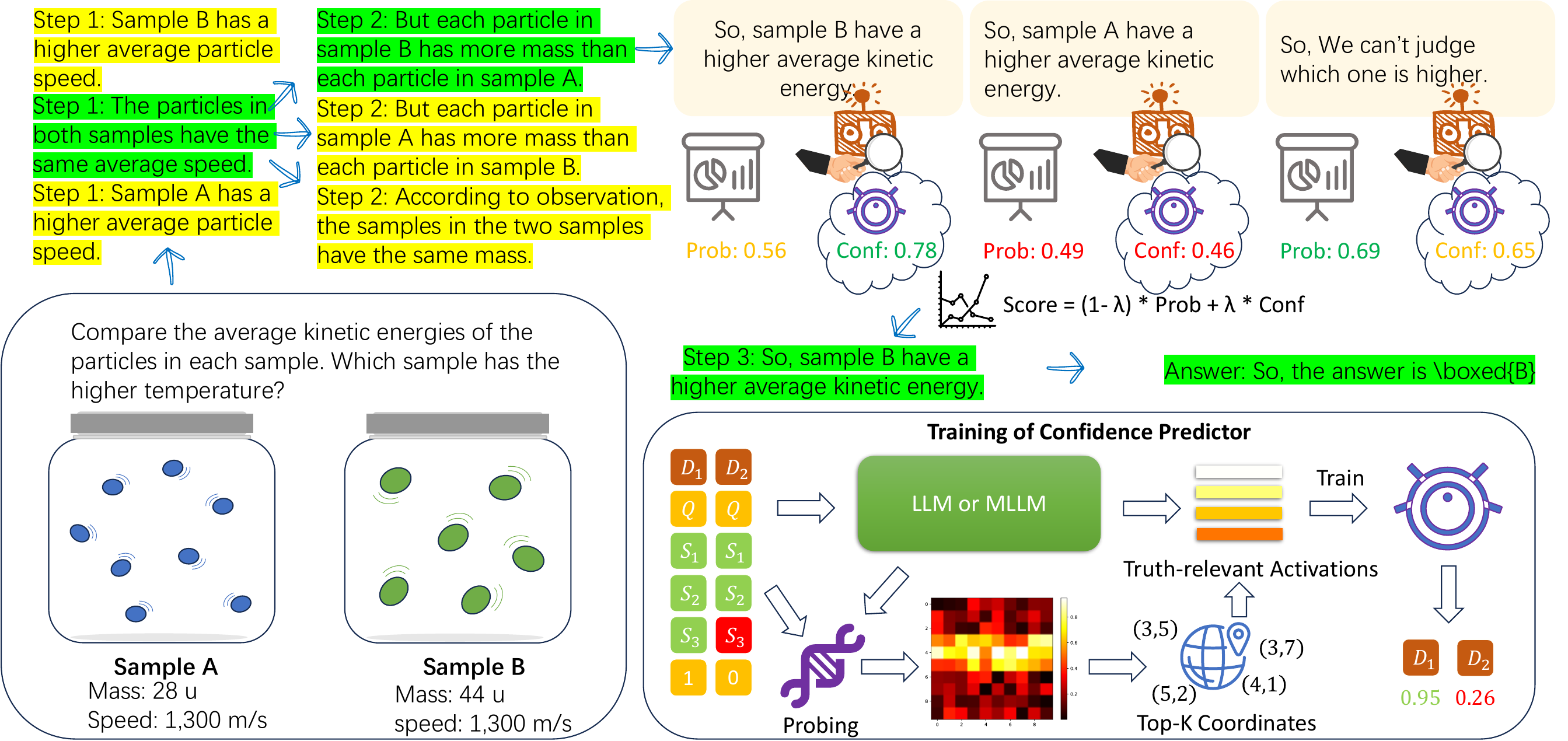}
    \caption{An overview of our method. Using the binary-labeled data to train a confidence predictor, and then introducing this predictor into the CoT reasoning process to select multiple generated candidates with the highest confidence.}
    \label{fig:flow}
\end{figure*}

\section{Related Work}
\subsection{CoT Decoding Strategy}

Chain-of-Thought (CoT) prompting \cite{wei2022chain} has revolutionized multi-step reasoning in LLMs by interleaving intermediate rationales into prompts. Early work demonstrated its efficacy across arithmetic, commonsense, and symbolic tasks, leveraging in-context learning to decompose problems into explicit reasoning chains. Subsequent advancements extended CoT to zero-shot settings \cite{kojima2022large} using generic instructions like ``Let’s think step by step,'' revealing LLMs' emergent reasoning capabilities. For structured tasks, few-shot CoT \cite{wei2022chain,zhang2022automatic} further improved performance through task-specific demonstrations, highlighting the importance of prompt complexity and relevance.
In multimodal contexts, adapted CoT frameworks \cite{wang2024t,zhang2023multimodal} often involve extracting image information as intermediate steps. For example, scene graph generation \cite{mitra2024compositional} and subtask decomposition \cite{gao2024cantor} enhance MLLM reasoning by incorporating visual context into sequential rationales.
To address error accumulation in long chains, UAG \cite{yin2024reasoning} identifies uncertain steps using token entropy and refines chains via certified clues. Techniques like ITI \cite{li2023inference} and DOLA \cite{chuang2023dola} further optimize reasoning by manipulating attention head activations or layer-specific logits. \citet{xie2024calibrating} introduced ``internal consistency" to calibrate LLM reasoning by measuring agreement between intermediate and final layer predictions.

\subsection{Uncertainty Quantification}
Uncertainty quantification, a cornerstone of machine learning, measures a model’s confidence in its predictions to ensure reliability. Typical approaches for deep learning includes MC dropout \cite{srivastava2014dropout}, Laplace approximation \cite{ritter2018scalable}, and deep ensembles \cite{lakshminarayanan2017simple}. In the era of LLMs, novel methods have emerged to address uncertainty in sequential generation. For example, predictive entropy \cite{kadavath2022language} quantifies uncertainty via logit distributions, while semantic entropy \cite{kuhn2023semantic,cui2020calibrated} captures semantic variability across multiple model responses. Recent work like ACTCAB \cite{liu2024enhancing} further refines this by training a linear layer on LM’s last-layer activations to get more precise confidence estimation.
Researchers have leveraged these techniques to detect hallucinations in LLMs \cite{farquhar2024detecting,zhang2024mathverse}, enhancing their trustworthiness. Notably, integrating uncertainty into CoT prompting has emerged as a promising direction. Methods like UAG \cite{yin2024reasoning} dynamically identify uncertain reasoning steps using token-level entropy. Similarly, \citet{mo2024tree} incorporates uncertainty-guided tree of thought to explore multiple reasoning paths. 

\section{Confidence from Truth-Relevant Attention Head Activations}

The research by \citet{li2023inference} made a important discovery: the internal activation patterns of LLMs implicitly encode the veracity of answers. Specifically, they used a labeled dataset containing both correct and incorrect answers, concatenated ``question-answer pairs'' as model inputs, and trained binary classification probes to analyze the sensitivity of each attention head to truthfulness. Their results revealed that attention heads in the middle layers of the Transformer exhibit the highest sensitivity to veracity signals. This finding provided the first empirical evidence of an implicit veracity judgment mechanism at the neural level, laying a critical foundation for understanding how LLMs ``reason'' about truth internally.

However, this framework is limited to static ``question + final answer'' paradigms and fails to capture the dynamic, step-by-step nature of CoT reasoning. In CoT processes, each intermediate reasoning step not only relies on its own logic but also depends on the reasoning context established by preceding steps. This dynamic characteristic raises a more complex question: 
Given a question and the generated preceding reasoning steps ($S_{1,\dots,n-1}$), do the model's internal activations contain implicit signals about the veracity of the ``next reasoning step'' ($S_n$)?

To address this question, we designed a binary classification dataset specifically tailored for CoT scenarios. Unlike the static question-answer data used by \citet{li2023inference}, each sample in our dataset captures the reasoning context of CoT: formatted as 
\((Q, S_{1,\dots,n-1}, S_{n}^{\text{true or false}})\)
where $Q$ is the original question, $S_{1,\dots,n-1}$ represents the generated preceding steps, $S_n$ is the next reasoning step to be evaluated (either true $S_{n}^{\text{true}}$ or false $S_{n}^{\text{false}}$), Appendix A for detailed dataset construction.

Using this dataset, we trained a targeted binary classification probe to test a core hypothesis: Can the model's internal activations effectively distinguish between truthful and false branches in CoT reasoning?  By measuring the classification accuracy of this probe on a validation set, we were able to quantify the sensitivity of activations in different layers and attention heads to the ``truthfulness of the next step''---this approach extends the probe analysis methodology of \citet{li2023inference} and for the first time explores implicit veracity signals in dynamic reasoning scenarios, providing an actionable framework for studying CoT reasoning.
To provide a better explanation, we introduce the following concepts:

\textbf{Multi-head attention architecture.}
To probe the internal activations of the model, we describe how to select the activation locations we need to probe within the Transformer architecture. A single transformer layer contains two key modules: a MHA mechanism and a standard multilayer perceptron (MLP) layer. During inference, the tokens are initially embedded into a high-dimensional space \( x_0 \in \mathbb{R}^{D_{high}} \), which serves as the starting point for the residual stream. This vector initiates the sequence \( x_0, x_1, \dots, x_n \) of vectors in the stream. Each transformer layer processes the current vector \( x_i \), performs computations, and adds the result to generate the next vector \( x_{i+1} \). The final vector is then decoded into a prediction for the next-token distribution. In each layer, the MHA consists of \( H \) separate linear transformations. Specifically, the MHA can be written as:

\begin{equation}
x_{l+1} = x_l + \sum_{h=1}^{H} Q_h^l Att^l (P_h^l x_l),
\end{equation}
where \( P_h^l \in \mathbb{R}^{D_{head} \times D_{high}} \) maps the activations into a head space, and \( Q_h^l \in \mathbb{R}^{D_{high} \times D_{head}} \) maps them back. The operator \( Att \) facilitates communication between input tokens. Following \citet{li2023inference}, the activations used for probe detection and training the confidence predictor occur after \( Att \) and before \( Q_h^l \), with activations denoted by \( x_h^l \in \mathbb{R}^{D_{head}} \). Specifically, these activations take the form:

\begin{equation}
\mathbf{x}_h^l = \text{Att} \left( P_h^l \mathbf{x}_l \right). \label{eq:AttentionHead}
\end{equation}

\textbf{Probing.}
A classification probe \cite{alain2018understanding} is a lightweight classifier used to examine whether a model’s internal representations encode specific discrete features. Its core idea is to model the mapping between internal activations and target labels in a formalized way, thereby quantifying the model’s implicit ability to encode features.

Typically, classification is performed using a linear transformation followed by a softmax function:
\(
f(h) = \text{softmax}(W^\top h + b)
\).
Given an input sample set \(
X =  \{x_1, x_2, \ldots, x_N\}
\),
we extract the activation $a$ from a specific head of the model:
\(
H = \{h_1, h_2, \ldots, h_N\}, \quad \text{where} \quad h_i = \text{Model}(x_i; \theta)^{(a)}
\)
Here, $\theta$ denotes the original model parameters, and $a$ is the target activation.
Using $H$ as input and the label set:
\(Y = \{y_1, y_2, \ldots, y_N\} \)
as supervision, we train the probe parameters $\{W, b\}$. In our experiment, we selected the hidden state of the last token in the sequence $X$.

After training, the classification accuracy of the probe on an independent validation set is used to quantify the encoding ability of the activations:
\(
\text{Acc} = \frac{1}{M} \sum_{j=1}^{M} \mathbf{1} \left( f(h_j^*) = y_j^* \right)
\)
where $M$ is the number of validation samples, and $(h_j^*, y_j^*)$ are the activation-label pairs from the validation set.
If the probe achieves accuracy significantly higher than random chance, it suggests that the corresponding layer's activations encode substantial information about the target feature.

\textbf{Interpretability case studies.} We conducted probing experiments on both unimodal and multimodal models. As illustrated in Figure \ref{fig:flow}, we used heatmaps to visualize the veracity classification accuracy of the trained probes on the validation set. It is evident that certain attention heads in the Transformer can distinguish between positive and negative CoT branch samples with remarkably high accuracy, reaching up to 85\%. This indicates that both unimodal and multimodal models possess an internal latent cognition capable of discerning the veracity of dynamically progressive CoT reasoning.
Consistent with the findings of \citet{li2023inference}, the activations sensitive to the veracity of CoT patterns are predominantly distributed in the middle layers. Notably, the distribution patterns of the multimodal LLaVA model and the unimodal LLaMA model exhibit a high degree of similarity. Given that LLaVA is trained on the LLaMA base model, this similarity suggests that such latent cognition may originate from pre-training rather than task-specific fine-tuning.

\begin{figure}[t]
    \centering
    \begin{subfigure}{\columnwidth}
        \centering
        \includegraphics[width=1\columnwidth]{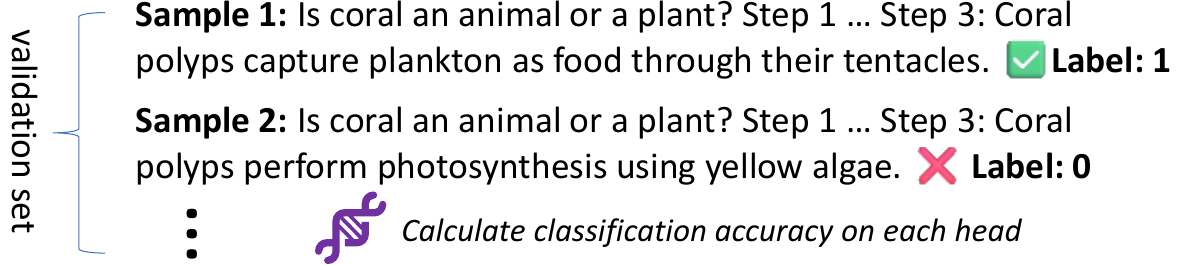}
        \label{fig:val_set}
    \end{subfigure}
    
    \vspace{0.5mm} 


    \caption{This figure shows the binary dataset of the constructed CoT.}
    \label{fig:heatmap_all}
\end{figure}


\subsection{Training of Confidence Predictor}
Merely identifying attention heads sensitive to truthfulness is insufficient to solve the problem. To transform these latent signals into operational tools, a rigorous quantification mechanism is required to convert distributed neural activations into interpretable confidence metrics. Specifically, we propose training a ``truthfulness-aware confidence predictor"—an auxiliary module that learns to map the model's internal states to a scalar value, which characterizes the prediction of the truthfulness of the next reasoning step. This module learns to associate the activation dynamics of sensitive heads with human-judged truthfulness.

Let \(x\) denote the input query, with \(y\) representing the model's output sequence of length \( |\mathbf{y}| \). For the final token \( y_{|\mathbf{y}|} \), we retrieve its hidden state \( \mathbf{h}^{l} \) from the \( l \)-th layer of the language model, which consists of \( H \) attention heads. Specifically, according to Equation \ref{eq:AttentionHead}, this state is decomposed into a multi-head representation:
\(
\mathbf{h}^{l} = \left[ \mathbf{h}^{l}_1, \mathbf{h}^{l}_2, \dots, \mathbf{h}^{l}_H \right] \in \mathbb{R}^{H \times D_{\text{hidden}}}
\).
where \( \mathbf{h}^{l}_h \in \mathbb{R}^{D_{\text{hidden}}} \) denotes the attention output for head \( h \). Unlike previous methods that aggregate activations across the entire sequence, our method focuses on the activations corresponding to the final token. This choice is motivated by the idea that the final token's representation encapsulates the model's judgment of the overall correctness of the generated sequence \cite{li2023inference}.

To identify which attention heads provide the most useful information for assessing truthfulness, we use a probing technique to analyze the correlation between each head's activation and the ground-truth label. This allows us to rank the attention heads by their relevance to truthfulness detection. We then select the top-\( K \) heads (\( K = H \), we choose to select the same number of attention heads as each layer) with the highest correlation, ensuring that we capture the most informative heads from the model’s intermediate layers:
\(
\mathcal{S} = \left\{ (l_1, h_1), (l_2, h_2), \dots, (l_K, h_K) \right\}
\).
Here, \( l_i \in \{1, 2, \dots, L\} \) and \( h_i \in \{1, 2, \dots, H\} \) denote the layer and head indices of the selected heads. 

Next, the activations of the selected heads are concatenated into a unified representation vector:
\begin{equation}
\mathbf{v} = \text{Concat}\left( \mathbf{h}^{l_1}_{h_1}, \mathbf{h}^{l_2}_{h_2}, \dots, \mathbf{h}^{l_K}_{h_K} \right) \in \mathbb{R}^{K \cdot D_{\text{hidden}}}.
\end{equation}
This concatenation captures the spatial dependencies encoded by each head while increasing the representational capacity of the feature vector.

To predict the confidence of the generated sequence $y$, we use a simple classifier:
\(
p_{\theta}(y|\mathbf{x}) = \sigma\left( \mathbf{W} \cdot \mathbf{v} + b \right)
\).
where \( \mathbf{W} \in \mathbb{R}^{1 \times (K \cdot D_{\text{hidden}})} \) and \( b \in \mathbb{R} \) are learnable parameters. The sigmoid function \( \sigma(\cdot) \) normalizes the output to a confidence score in \([0,1]\). This classifier explicitly models the relationship between the attention head activations and the model's confidence, enabling a more reliable estimation of the output's correctness.

\textbf{ECE Loss.} To train the confidence predictor, the standard mean squared error (MSE) loss function is commonly applied in calibration tasks. However, it inherits a fundamental limitation from binary labels,  This sharp \(0/1\) target often leads to overconfidence or underconfidence. \cite{guo2017calibration}.


To overcome this limitation, we adopt the ``ECE loss" proposed by \citet{liu2024enhancing} to improve calibration. They demonstrated its effectiveness through ablation experiments. 

The ECE loss replaces binary correctness labels with soft targets derived from cross-validation accuracy distributions. Specifically, \(K\)-fold cross-validation is used to generate confidence scores for each fold, and the data is partitioned into \(B\) equal-interval bins. For each bin \(B_i\), the empirical accuracy is computed as:
\(
\text{acc}(B_i) = \frac{1}{|B_i|} \sum_{(x,y) \in B_i} \mathbf{1}(y \text{ is correct}),
\)
where \( \text{acc}(B_i) \) represents the proportion of correct predictions in the bin \( B_i \). These accuracies provide a soft label that better reflects the true confidence-accuracy relationship. The final ECE loss is then computed as:
\begin{equation}
\mathcal{L}_{\text{ECE}} = \frac{1}{|\mathcal{D}|} \sum_{(x,y) \in \mathcal{D}} \left( \text{acc}(B_i) - p_{\theta}(y|x) \right)^2.
\end{equation}
This loss function directly aligns the predicted confidence with the empirical accuracy of the model's predictions, leading to more accurate calibration of the confidence scores.

\subsection{Confidence Predictor Evaluation}
Given the limited persuasiveness of comparing the performance of the trained predictor with other baseline methods when evaluating it on intermediate steps of CoT reasoning, we ultimately opted to use a non-CoT dataset to assess its calibration for predicting the truthfulness of final answers. The construction of the dataset required for predictor training is detailed in Appendix A.

\textbf{Model.} We tested our approach using LLaMA2-7B-Chat, LLaMA2-13B-Chat, and LLaMA-3-8B-Instruct models.

\textbf{Metrics.} We employ three complementary metrics to evaluate calibration and prediction quality: Expected Calibration Error (ECE), Brier Score and Area Under the ROC Curve (AUC). See Appendix B for specific definitions.

A model may achieve high AUC yet exhibit poor calibration (high ECE/Brier Score), underscoring the necessity of evaluating all three metrics to fully characterize predictive reliability.

\textbf{Baseline.} We compared four methods for quantifying confidence to demonstrate the superiority of our approach.
\begin{itemize}
    \item \textbf{Sequence Likelihood:} The average probability value assigned by the model to the answer tokens.
    \item \textbf{``Is True'' probability:} The probability assigned by the language model to ``True'' when prompted to verify the correctness of a question-answer pair \cite{tian2023just}. 
    \item \textbf{Verbalization:} Method where the language model is prompted to verbally articulate its confidence of answers \cite{tian2023just}.
    \item \textbf{ACTCAB:} Method that exclusively uses the final layer of the model to train a confidence predictor \cite{liu2024enhancing}.
\end{itemize}

\begin{figure}[t]  
    \centering
    \includegraphics[width=0.9\columnwidth]{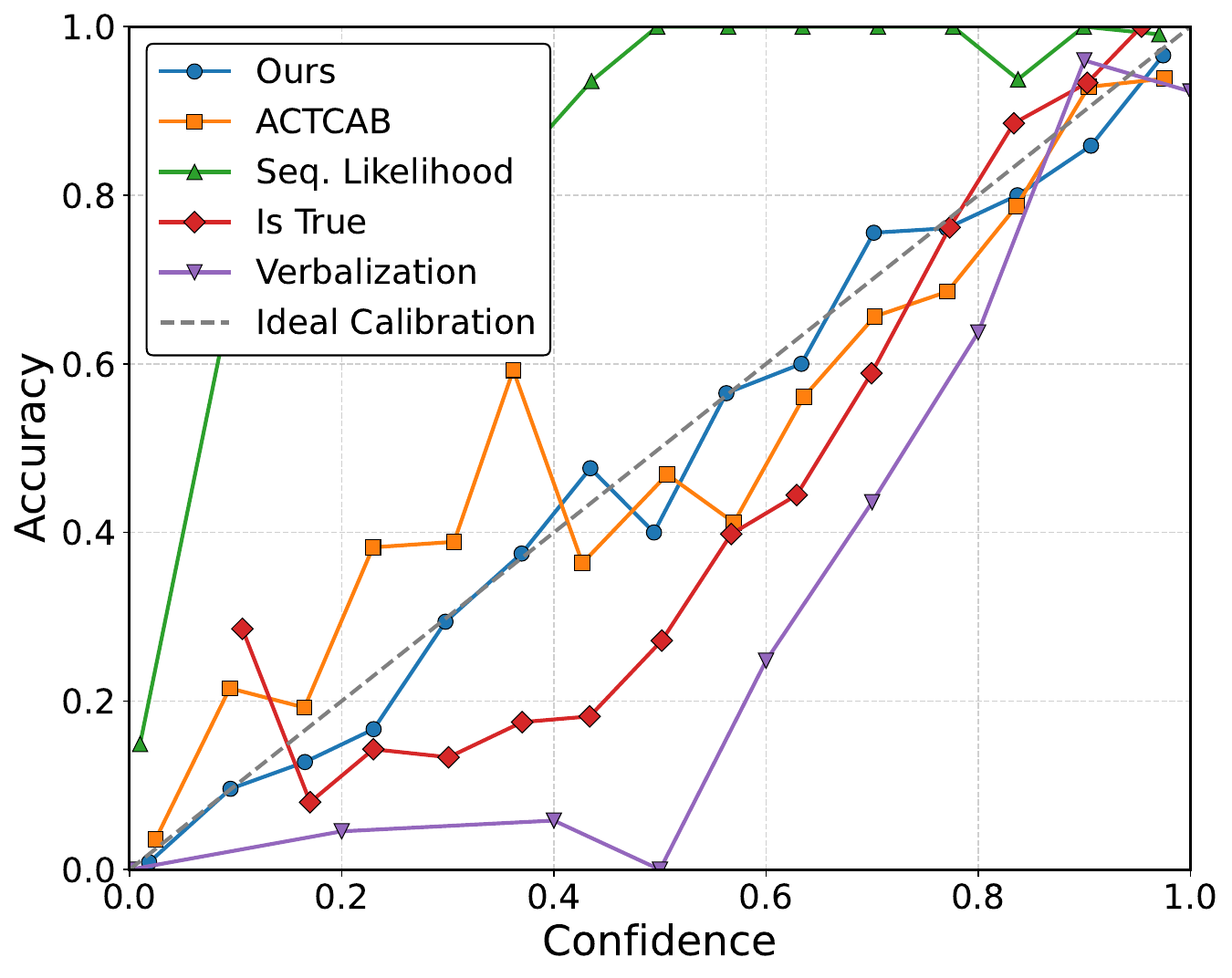}
    \caption{This figure shows the calibration curves of LLaMA2-13B-Chat in SciQ, where closer proximity to the Ideal Calibration curve indicates better calibration.}
    \label{fig:uc_curve}
\end{figure}

\textbf{Results.} To evaluate the effectiveness of our method, we compared it against four confidence quantification approaches on the test sets of four non-CoT datasets: CommonsenseQA \cite{talmor-etal-2019-commonsenseqa}, SciQ \cite{SciQ}, WikiQA \cite{yang-etal-2015-wikiqa}, and TruthfulQA \cite{lin2021truthfulqa}. Table \ref{tab:confidence_eval} presents the results (The remaining results are in Appendix B).
Notably, ACTCAB and our proposed method (both trained approaches) demonstrated substantial performance gaps over untrained baselines. Specifically, our approach, which incorporates Truth-Relevant attention activations identified via probing, outperformed ACTCAB’s exclusive reliance on the model’s final layer for confidence prediction. Additionally, the use of ECE loss optimized both ACTCAB and our method for improved calibration, reflected in their strong ECE scores. While Brier Score and AUC provided broader assessments beyond calibration-specific optimization, our method consistently outperformed competitors across datasets, models, and metrics. Figure \ref{fig:uc_curve} displays a visual comparison of calibration curves.

\begin{table}[htbp]
\scriptsize
\setlength{\tabcolsep}{3pt} 
\renewcommand{\arraystretch}{0.7}
\centering
\begin{tabular}{l@{\hskip 0.5em}l@{\hskip 0.6em}ccccc}
\toprule
Dataset & Metric & Ours & ACTCAB & Seq. Likelihood & Is True & Verbalization \\
\midrule[1.2pt]

\multirow{3}{*}{WikiQA}
  & ECE$\downarrow$   & \textbf{0.037} & 0.058 & 0.271 & 0.146 & 0.254 \\
  & Brier$\downarrow$ & \textbf{0.102} & 0.149 & 0.302 & 0.231 & 0.291 \\
  & AUC$\uparrow$     & \textbf{0.934} & 0.868 & 0.678 & 0.747 & 0.640 \\

\midrule[0.8pt]
\multirow{3}{*}{TruthfulQA}
  & ECE$\downarrow$   & \textbf{0.019} & 0.038 & 0.284 & 0.155 & 0.178 \\
  & Brier$\downarrow$ & \textbf{0.122} & 0.140 & 0.344 & 0.286 & 0.296 \\
  & AUC$\uparrow$     & \textbf{0.907} & 0.881 & 0.480 & 0.510 & 0.570 \\

\midrule[0.8pt]
\multirow{3}{*}{SciQ}
  & ECE$\downarrow$   & \textbf{0.018} & 0.052 & 0.223 & 0.132 & 0.194 \\
  & Brier$\downarrow$ & \textbf{0.095} & 0.132 & 0.183 & 0.178 & 0.221 \\
  & AUC$\uparrow$     & 0.937 & 0.897 & \textbf{0.949} & 0.848 & 0.784 \\

\midrule[0.8pt]
\multirow{3}{*}{CommonQA}
  & ECE$\downarrow$   & \textbf{0.024} & 0.029 & 0.167 & 0.482 & 0.224 \\
  & Brier$\downarrow$ & \textbf{0.133} & 0.137 & 0.184 & 0.481 & 0.290 \\
  & AUC$\uparrow$     & \textbf{0.891} & 0.885 & 0.873 & 0.422 & 0.604 \\

\bottomrule[1.2pt]
\end{tabular}
\caption{Evaluation of our confidence predictor against state-of-the-art baselines on LLaMA2-13B-Chat.}
\label{tab:confidence_eval}
\end{table}

\section{More Reliable Reasoning Chain Generation} \label{sec:cot}
A single reasoning path is often constrained by local optima or implicit biases, making it difficult to fully unleash the model’s internal knowledge reserves. Therefore, exploring multiple potential reasoning branches through multi-path sampling has emerged as a key approach to excavate the model’s deep-seated knowledge and mitigate the risks of a single path. To this end, we integrate the well-trained confidence predictor into the CoT generation process: it quantitatively evaluates the next reasoning step of each candidate path, ultimately selecting the step with the highest confidence as the cornerstone for subsequent reasoning.

\subsection{Confidence Predictor Guided Beam Search}
We propose an improved sequence generation method that optimizes CoT reasoning through dynamic fusion of confidence predictions and generation probabilities. The core workflow can be summarized as follows: at each decoding step, beam search generates multiple candidate reasoning steps, then a pretrained confidence predictor evaluates their quality, and a weighted scoring function ultimately selects the optimal path. Specifically, given the current generated chain-of-thought sequence $ S_t = \{s_1, s_2, \dots, s_t\} $, where each $ s_i $ represents a reasoning step, the model extends it by generating $ M $ candidate subsequent steps $ \{C_{t+1}^1, C_{t+1}^2, \dots, C_{t+1}^M\} $ through beam search. Each candidate step $ C_{t+1}^m $ consists of a sequence of tokens $ (x_1^m, x_2^m, \dots, x_{L_m}^m) $, generated with probability distribution $ P(C_{t+1}^m | S_t) $. The pretrained confidence predictor $ \mathcal{F}_{\text{conf}} $ computes a confidence score $ \beta(C_{t+1}^m) $ for each candidate. These scores are combined with the normalized token-level generation probability to construct a global scoring function:

\[
\bar{P}(C_{t+1}^m) = \exp(
\frac{1}{L_m} \sum_{k=1}^{L_m} \log P(x_k^m | S_t, x_1^m, \dots, x_{k-1}^m)
),
\]
\begin{equation}
\text{Score}(C_{t+1}^m) = \lambda \cdot \beta(C_{t+1}^m) + (1 - \lambda) \cdot \bar{P}(C_{t+1}^m),
\label{eq:score}
\end{equation}

where $ \lambda \in [0, 1] $ is a balancing weight (we performed grid search on several datasets and ultimately selected $\lambda = 0.5$,) and $ L_m $ denotes the number of tokens in the candidate step $ C_{t+1}^m $. The algorithm appends the highest scoring candidate $ C_{t+1}^* $ to the current chain $ S_t $, forming $ S_{t+1} = S_t \oplus C_{t+1}^* $, and iterates until the final answer is generated. By guiding the search direction with the confidence predictor, our method mitigates the degradation of the reasoning path caused by the accumulation of errors. Detail settings are in Appendix C. Figure \ref{fig:flow} shows the overall process of our method.

\subsection{Experiments}
We systematically evaluate the performance of our proposed method by leveraging state-of-the-art LLM and MLLM backbones across a diverse range of reasoning benchmarks.

\textbf{Benchmarks.}
The evaluation framework comprises two primary categories: \emph{Mathematical/Symbolic Reasoning} and \emph{Commonsense Reasoning}, each spanning both unimodal and multimodal scenarios.

\begin{itemize}
\item \textbf{Mathematical \& Symbolic Reasoning}
    The unimodal benchmarks include \emph{GSM8K} \cite{cobbe2021gsm8k}, \emph{Boolean\_Expressions} \cite{suzgun2022challenging} and \emph{SVAMP} \cite{patel2021nlp}. Multimodal evaluation is represented by \emph{CLEVR-Math} \cite{lindstrom2022clevr}.
    \item \textbf{Commonsense Reasoning}
    Unimodal evaluation includes \emph{StrategyQA} \cite{geva2021did} and \emph{BoolQ} \cite{clark2019boolq}. Multimodal benchmarks include \emph{ScienceQA} \cite{lu2022learn}, \emph{RealWorldQA} \cite{RealWorldQA}, and \emph{MMStar} \cite{chen2024we}.
\end{itemize}

\textbf{Models.}
To comprehensively evaluate our method, we conducted experiments using state-of-the-art LLMs and MLLMs spanning diverse parameter scales and architectures. For unimodal reasoning, we employed LLaMA-2 in 7B, 13B, and 70B \cite{touvron2023llama} configurations to examine scalability effects. In the multi-modal domain, we utilized LLaVA v1.6 (7B/13B, \cite{liu2023visual}) and Qwen2.5-VL-Instruct (7B, \cite{qwen2.5-VL}) to assess cross-modal capabilities. 

\textbf{Baseline.}
To conduct a comprehensive comparison of our method, we selected three representative baselines:
(1) Few-Shot CoT \cite{wei2022chain}, this approach provides a small number of reasoning examples to prompt the model for step-by-step reasoning;
(2) Self-Consistency \cite{wang2022self}, it generates multiple reasoning paths through sampling and selects the optimal chain via majority voting;
(3) Self-Evaluation Guided Beam Search \cite{xie2023self}, for fair comparison, it shares identical hyperparameters and prompt templates with our proposed approach, except replacing our scoring function (Equation \ref{eq:score}) with their metric.
(4) Process Reward Models \cite{zhao2025genprm} aim to score each step of the reasoning process and guide model learning through fine-grained supervision. PRMs replace the models with a classification head to provide reward scores.

\begin{table*}[htbp]
\scriptsize
\setlength{\tabcolsep}{3pt} 
\renewcommand{\arraystretch}{0.7} 
\centering
\begin{minipage}[t]{0.48\textwidth}
\centering
\begin{tabular}{@{}lccccccc@{}}
\toprule
\multirow{2}{*}{Model} & \multirow{2}{*}{Method} & \multicolumn{5}{c}{Datasets} & \multirow{2}{*}{AVG} \\
\cmidrule{3-7}
 & & GSM8K & SVAMP & StrategyQA & BoolQ & Boolean & \\
\midrule
\multirow{4}{*}{LLaMA2-7B} 
 & CoT-few & 24.4 & 43.3 & 63.0 & 54.8 & 62.4 & 49.6 \\
 & SC & 24.3 & 47.7 & 63.0 & 58.0 & 70.4 & 52.7 \\
 & SE & 24.9 & 40.0 & 63.0 & 58.2 & \textbf{72.0} & 51.6 \\
 & PRM & 24.4 & 44.0 & 63.8 & 56.6 & 67.2 & 51.2 \\
 & Ours & \textbf{25.2} & \textbf{48.3} & \textbf{64.5} & \textbf{58.6} & 70.0 & \textbf{53.3} \\
\cmidrule(lr){2-8}
\multirow{4}{*}{LLaMA2-13B} 
 & CoT-few & 39.9 & 53.7 & 66.0 & 57.6 & 68.8 & 57.2 \\
 & SC & 39.3 & 54.0 & 65.9 & 56.6 & \textbf{70.0} & 57.2 \\
 & SE & 38.4 & 51.7 & 66.7 & 56.4 & 69.6 & 56.6 \\
 & PRM & 39.2 & 55.0 & 65.8 & 53.8 & 68 & 56.4 \\
 & Ours & \textbf{42.8} & \textbf{55.7} & \textbf{66.8} & \textbf{59.2} & 68.8 & \textbf{58.7} \\
\cmidrule(lr){2-8}
\multirow{4}{*}{LLaMA2-70B} 
 & CoT-few & 52.3 & 70.3 & 72.0 & 67.0 & 79.6 & 68.2 \\
 & SC & \textbf{59.0} & 73.0 & 74.2 & 67.4 & 79.2 & 70.6 \\
 & SE & 52.6 & 66.3 & 74.3& 68.6 & 79.6 & 68.3 \\
 & PRM & 54.6 & 70.3 & \textbf{74.4} & 68.2 & 81 & 69.9 \\
 & Ours & 58.3 & \textbf{74.7} & 73.7 & \textbf{69.8} & \textbf{81.2} & \textbf{71.5} \\
\bottomrule
\end{tabular}
\end{minipage}
\hfill
\begin{minipage}[t]{0.48\textwidth}
\centering
\begin{tabular}{@{}lcccccc@{}}
\toprule
\multirow{2}{*}{Model} & \multirow{2}{*}{Method} & \multicolumn{4}{c}{Datasets} & \multirow{2}{*}{AVG} \\
\cmidrule{3-6}
 & & ScienceQA & realworldqa & clevr-math & MMStar & \\
\midrule
\multirow{4}{*}{LLaVA-7B} 
 & CoT-few & 58.3 & 23.0 & \textbf{18.0} & 40.3 & 34.9 \\
 & SC & 55.7 & 18.8 & 15.7 & 41.3 & 32.9 \\
 & SE & 61.6 & 28.2 & \textbf{18.0} & 44.0 & 38.0 \\
 & PRM & \textbf{65.0} & 33.1 & 17.0 & 44.0 & 39.7 \\
 & Ours & 62.4 & \textbf{33.7} & \textbf{18.0} & \textbf{45.7} & \textbf{40.0} \\
\cmidrule(lr){2-7}
\multirow{4}{*}{LLaVA-13B} 
 & CoT-few & 61.9 & 31.9 & 10.7 & 41.0 & 36.4 \\
 & SC & 64.0 & 25.6 & 13.0 & 39.3 & 35.5 \\
 & SE & 65.4 & 30.3 & 11.7 & 41.7 & 37.3 \\
 & PRM & 61.3 & \textbf{33.9} & 9.7 & 41.3 & 36.6 \\
 & Ours & \textbf{69.2} & 31.9 & \textbf{14.0} & \textbf{42.0} & \textbf{39.3} \\
\cmidrule(lr){2-7}
\multirow{4}{*}{Qwen2.5-VL-7B} 
 & CoT-few & \textbf{86.0} & 52.2 & 64.7 & 58.0 & 65.2 \\
 & SC & 84.0 & 52.7 & 67.7 & 57.3 & 65.4 \\
 & SE & 76.3 & \textbf{53.7} & 68.5 & 57.7 & 64.1 \\
 & PRM & 81.0 & 53.0 & 70.0 & 58.1 & 65.5 \\
 & Ours & 82.0 & 51.2 & \textbf{71.0} & \textbf{58.3} & \textbf{65.6} \\
\bottomrule
\end{tabular}
\end{minipage}
\caption{Comparison of Chain-of-Thought Decoding Processes in Unimodal (left) and Multimodal (right) tasks. SC: Self-Consistency, SE: Self-Evaluation Guided Beam Search.}
\label{tab:cot_comparison_both}
\end{table*}

\subsection{Main Results}
Table \ref{tab:cot_comparison_both} present the performance of our proposed method in unimodal and multimodal tasks respectively. The results clearly show that our method achieves excellent performance under each setting. When compared with the few - shot CoT under the same settings, our method demonstrates significant improvements in most tests. For instance, in the unimodal task, for the SVAMP dataset, our method has a 5\% improvement over the few shot CoT (48.3 vs. 43.3). In the multimodal task, for the realworldqa dataset, our method shows a 10.7\% improvement. Overall, our method outperforms the baseline few - shot CoT in most cases, whether it is in Mathematical \& Symbolic Reasoning, Commonsense Reasoning tasks, or unimodal and multimodal tasks. This fully demonstrates that the confidence extracted from the internal states of the model can effectively guide the generation of more reliable reasoning chains. Some examples shown in Appendix C.

\begin{table}[htbp]
\scriptsize
\setlength{\tabcolsep}{3pt} 
\renewcommand{\arraystretch}{0.9} 
\centering
\begin{tabular}{lccccccc}
\toprule
\multirow{2}{*}{Model} & \multirow{2}{*}{Method} & \multicolumn{5}{c}{Datasets} & \multirow{2}{*}{AVG} \\
\cmidrule{3-7}
 & & GSM8K & SVAMP & StrategyQA & BoolQ & Boolean & \\
\midrule

\multirow{4}{*}{DS-R1-7B} 
 & CoT-few & 85.3 & 87.3 & 56.9 & 56.6 & 90.4 & 75.3 \\
 & SC & 84.2 & 87.5 & 60.2 & 56.6 & 95.2 & 76.7 \\
 & SE & 82.3 & 85.7 & 59.8 & \textbf{59.8} & 96 & 76.7 \\
 & PRM & \textbf{86.2} & 88.2 & \textbf{60.7} & 57.4 & 95.2 & 77.5 \\
 & Ours & 84.0 & \textbf{88.3} & 59.7 & 59.4 & \textbf{96.8} & \textbf{77.6} \\

\bottomrule
\end{tabular}
\caption{This table demonstrates the performance of our approach on the DeepSeek R1 distilled reasoning model.}
\label{tab:deepseek}
\end{table}
\subsection{The Impact of Confidence Selection}

\emph{It is important to rely on confidence predictors for selection among candidates.}
To verify the effect of the confidence predictor we introduced on the results, we conducted an ablation study. Instead of selecting candidates with higher confidence based on Equation \ref{eq:score}, we randomly selected from these candidates. With all other parameter settings unchanged, we tested on multiple datasets and compared few-shot CoT, our method, and the random selection strategy, as shown in Figure \ref{fig:random13B}. The results indicate that (the remaining results are in Appendix D), compared to when the confidence predictor is used, the random strategy shows a significant drop in performance, and in many cases, its performance is even lower than the baseline few-shot CoT. This ablation study strongly demonstrates the importance of the confidence predictor.

\subsection{Effectiveness of Self-Correcting Error Steps}

\begin{figure}[t]  
    \centering
    \includegraphics[width=1\columnwidth]{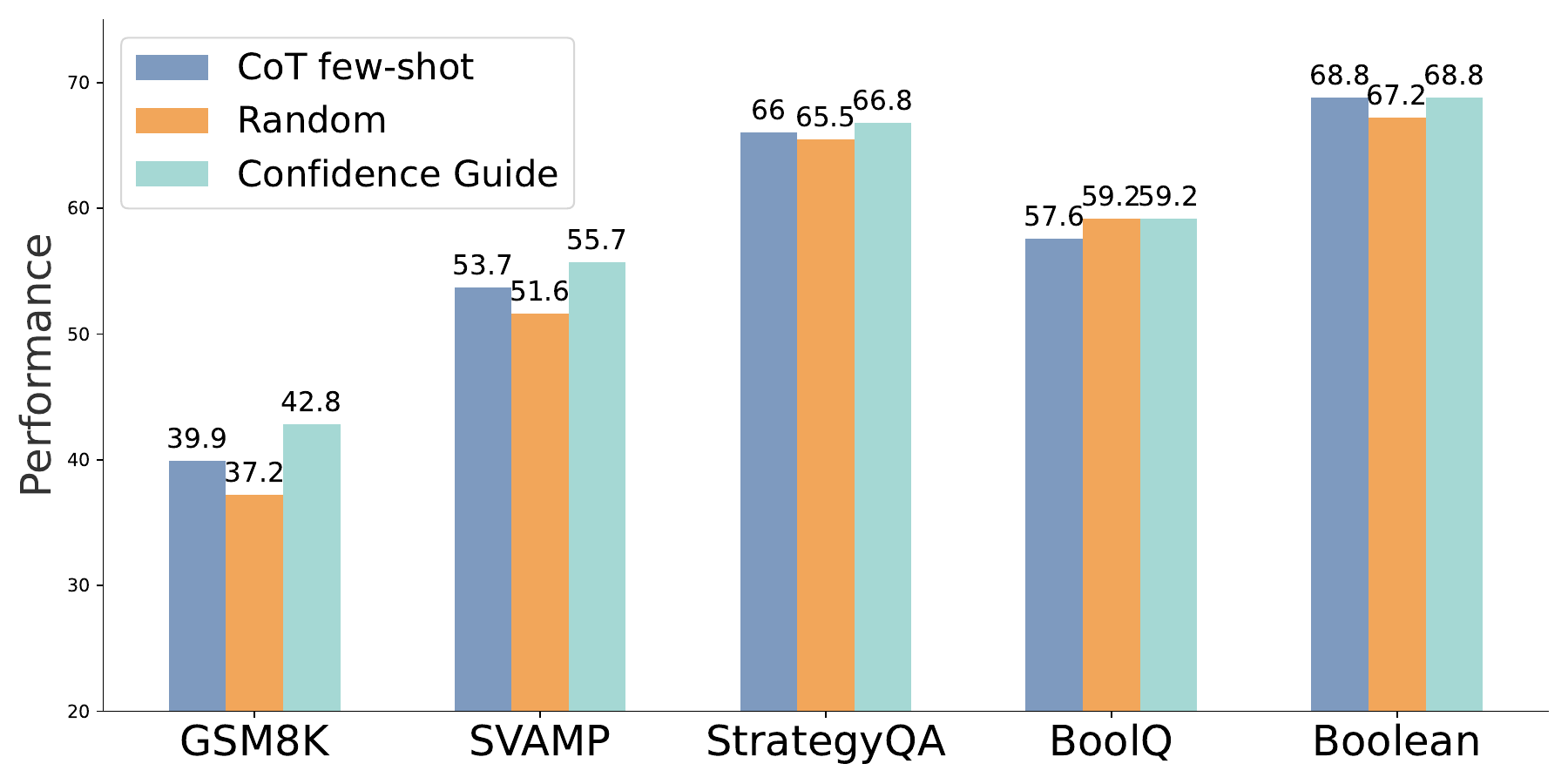}
    \caption{This figure demonstrates the effect (LLaMA2-13B) of confidence-guided reasoning.}
    \label{fig:random13B}
\end{figure}


\begin{table}[htbp]
\scriptsize
  \centering
  \begin{tabular}{lrrrrrr}
    \toprule
    Method & \multicolumn{1}{l}{GSM8K} & \multicolumn{1}{l}{SVAMP} & \multicolumn{1}{l}{StrategyQA} & \multicolumn{1}{l}{BoolQ} & \multicolumn{1}{l}{Boolean} & \multicolumn{1}{l}{AVG} \\
    \midrule
    Ours & 25.2 & 48.3 & \textbf{64.5} & 58.6 & \textbf{70.0} & 53.3 \\
    Revise    & \textbf{26.0} & \textbf{49.0} & \textbf{64.5} & \textbf{59.6} & 69.6 & \textbf{53.7} \\
    \bottomrule
  \end{tabular}
  \caption{The effect of model (LLaMA2-7B) self-correction on erroneous steps in CoT reasoning.}
  \label{tab:revise7B}
\end{table}

\emph{Overall effective, indicating the compatibility of our method.}
In selecting the highest confidence candidate, low confidence levels across all candidates can lead to error accumulation in the CoT process. An intuitive solution is to set a threshold: if the confidence falls below it, the model self-corrects the potentially erroneous step. 
Some research suggests that unbiased prompts—those not explicitly instructing the model to correct errors—can encourage limited self-correction \cite{liu2024large}, while overly directive prompts may lead to blind adherence to instructions rather than reasoning.
To test this, we applied the unbiased prompting method from \citet{liu2024large} (see Appendix D) and set a threshold of $0.5$. The results show an overall improvement (see Table \ref{tab:revise7B}). These findings indicate that self-correction during CoT generation is effective, and our method is well compatible with some CoT error correction techniques.

\subsection{Performance in LRM}

\emph{Still effective, indicating the generalization capability of our method.}
We aim to investigate the applicability of our method to reasoning models like DeepSeek R1, which have undergone specialized CoT training, including reinforcement learning on long-chain reasoning datasets. We tested language models distilled from DeepSeek R1, such as DS-Distill-Qwen-7/14B.
Our results, shown in Table \ref{tab:deepseek} (full details in Appendix D), the overall performance still favors our method. 

\section{Conclusion}
This paper presents a method leveraging the internal truthfulness sensitivity of LLMs and MLLMs to significantly enhance the reliability of CoT reasoning. By analyzing attention head activations, we train a confidence predictor to evaluate the correctness of each reasoning step and design a confidence-driven reasoning generation strategy that effectively reduces error accumulation. Experimental results demonstrate superior performance across diverse tasks and model scales, outperforming existing approaches. Further analysis reveals the model's self-correction mechanism in CoT reasoning.

\section{Acknowledgments}
This paper is supported by the National Science Foundation of China Project (No. 62306098), Fundamental Research Funds for the Central Universities (No. JZ2024HGTB0256) and the Open Project of Anhui Provincial Key Laboratory of Multimodal Cognitive Computation, Anhui University (No. MMC202412). The computation is completed on the HPC Platform of Hefei University of Technology.

\bibliography{aaai2026}

\end{document}